\documentclass{svproc}

\usepackage{url}
\usepackage{amsmath,amssymb,amsfonts}
\usepackage{algorithmic}
\usepackage{graphicx}
\usepackage{multirow}

\usepackage{perpage}
\usepackage{xspace}
\usepackage{xcolor}
\usepackage{graphicx} 
\usepackage{adjustbox} 
\usepackage{caption}
\usepackage{booktabs}
\usepackage{cases}
\MakePerPage{footnote} 
\usepackage{hyperref}
\usepackage{rotating}

\newcommand{\CaseName}[1]{\textbf{[Case Name: #1]}}
\newcommand{\Action}[1]{\textbf{[Action: #1]}}
\newcommand{\Location}[1]{\textbf{[Location: #1]}}
\newcommand{\LegalConcept}[1]{\textbf{[Legal Concept: #1]}}
\newcommand{\Outcome}[1]{\textbf{[Outcome: #1]}}

\newcommand{\bert}{BERT\xspace}
\newcommand{\lbert}{LegalBERT\xspace}
\newcommand{\abert}{AnglEBERT\xspace}

\begin{document}
\mainmatter

\title{CBR-RAG: Case-Based Reasoning for Retrieval Augmented Generation in LLMs for Legal Question Answering \thanks{This research is funded by SFC International Science Partnerships Fund. }}
\titlerunning{CBR for RAG in LLMs for Legal QA} 

\author
{Nirmalie Wiratunga\inst{1}\and 
Ramitha Abeyratne\inst{1} \and 
Lasal Jayawardena\inst{1,2} \and
Kyle Martin\inst{1} \and
Stewart Massie\inst{1} \and
Ikechukwu Nkisi-Orji\inst{1} \and 
Ruvan Weerasinghe\inst{2}\and 
Anne Liret\inst{3} \and 
Bruno Fleisch\inst{3} }
\authorrunning{N. Wiratunga et al.}

\institute{Robert Gordon University, Aberdeen, UK\\
\email{\{n.wiratunga, r.abeyratne, l.jayawardena, k.martin3, s.massie, i.nkisi-orji\}@rgu.ac.uk},
\and
Informatics Institute of Technology, Sri Lanka\\
\email{ruvan.w@iit.lk}
\and
BT France\\
\email{\{anne.liret, bruno.fleisch\}@bt.fr}
}

\maketitle

\begin{abstract}
Retrieval-Augmented Generation (RAG) enhances Large Language Model (LLM) output by providing prior knowledge as context to input. This is beneficial for knowledge-intensive and expert reliant tasks, including legal question-answering, which require evidence to validate generated text outputs. We highlight that Case-Based Reasoning (CBR) presents key opportunities to structure retrieval as part of the RAG process in an LLM. 
We introduce CBR-RAG, where CBR cycle's initial retrieval stage, its indexing vocabulary, and similarity knowledge containers are used to enhance LLM queries with contextually relevant cases. This integration augments the original LLM query, providing a richer prompt.
We present an evaluation of CBR-RAG, and examine different representations (i.e. general and domain-specific embeddings) and methods of comparison (i.e. inter, intra and hybrid similarity) on the task of legal question-answering. Our results indicate that the context provided by CBR's case reuse enforces similarity between relevant components of the questions and the evidence base leading to significant improvements in the quality of generated answers.

\keywords{CBR, RAG, LLMs, Text Embedding, Indexing, Retrieval}
\end{abstract}

\section{Introduction}
Retrieval-Augmented Generation (RAG) enhances the performance of large language models (LLMs) on knowledge-intensive NLP tasks by combining the strengths of pre-trained parametric (language models with learned parameters) and non-parametric (external knowledge resources e.g., Wikipedia) memories~\cite{lewis2020retrieval}. This hybrid approach not only sets new benchmarks in open-domain question answering by generating more accurate, specific, and factually correct responses but also addresses critical challenges in the field such as the difficulty in updating stored knowledge and providing provenance for generated outputs. For example, in the context of legal question-answering, RAG-based systems can retrieve publicly available legislation documents from open knowledge bases to provide context for user queries. However, such a system would require output that could be validated, and moderation of content generated by LLMs has been highlighted as a critical concern~\cite{hacker2023regulating}.


Case-Based Reasoning (CBR) presents an excellent opportunity here, as previous solutions form the basis of a knowledge-base which can be evidenced in best practice or regulations~\cite{rissland1995hybrid}.
CBR can enhance the retrieval process in RAG models by organising the non-parametric memory in a way that cases (knowledge entries or past experiences) are more effectively matched to queries. 
Previous work on ensemble CBR-neural systems has also highlighted the benefits of CBR integration, demonstrating improvements in factual correctness over purely neural methods~\cite{upadhyay2022case}.
Accordingly, in this work we present three contributions. 
\begin{itemize}

\item Firstly, we formalise the role of the CBR methodology to form context for RAG systems. \item Secondly, we provide an empirical comparison of different retrieval methods for RAG, examining different  representations (i.e. general and domain-specific embeddings) and methods of comparison (i.e. inter, intra and hybrid similarity). 
\item Finally, we present these contributions in the context of the legal domain, and provide results for a generative legal question-answering (QA) application \footnote{Reproducible code is available: \url{https://github.com/rgu-iit-bt/cbr-for-legal-rag}}. Our results highlight the opportunities of CBR-RAG systems for knowledge-reliant generative tasks.
\end{itemize}

This paper is structured as follows. In Section~\ref{sec:related} we describe influential work targeting the legal domain from CBR and LLM literature. In Section~\ref{sec:cbrrag}, we formalise CBR-RAG and its application to legal QA, while in Section~\ref{sec:indexing} we detail the different methods for creating and comparing embeddings to perform case retrieval. In Section~\ref{sec:encoders} we describe the different encoders used to create embeddings. Finally, in Section~\ref{sec:eval} we discuss the methodology and results of our evaluation, followed by conclusions in  Section~\ref{sec:conc}.

\section{Related Work}
\label{sec:related}
CBR boasts a long-standing history in the legal domain, with initial efforts concentrating on extracting features to index law cases effectively. These features,  referred to as `factors', are pivotal in systems like HYPO~\cite{ashley1991reasoning}, which employ fact-oriented representations applied to trade secret law and extended to legal tutoring with the CATO~\cite{aleven1997teaching} system.
More generally with Textual CBR, a key focus has been on extracting features for case comparison, using methods that range from decision trees, as exemplified by the SMILE system in creating indexing vocabularies for legal case retrieval~\cite{bruninghaus2001role}, to association rules for case representation~\cite{wiratunga2004feature}. Much of this has now been advanced by the adoption of neural transformations of input text through transformer-style embeddings with LLMs.

The application of LLMs presents an interesting approach in addressing challenges within the legal domain. LLMs use language understanding capabilities to interact with users, enabling  extraction of key elements from legal documents, and present information in an understandable manner to enhance decision-making processes. 
For this reason, GPT-based~\cite{lee2023lexgpt,tang2023policygpt} and \bert-based~\cite{chalkidis2020legalbert} transformer models are popular for LegalAI. 
However their black-box nature and tendency to hallucinate and lack of factual faithfulness present significant challenges for deployment~\cite{lailaw2023large}.
Retrieval-Augmented Generation (RAG) systems address this by presenting the LLM with factual data to generate responses~\cite{lewis2020retrieval,li2023angleoptimized}, employing a variety of sophisticated fact identifying mechanisms~\cite{asai2023selfrag,thulke2021efficient}.
However currently such retrieval methods in RAG do not make use of CBR's potential for varying matching strategies across different segments of the content being matched. 
Here we are reminded of research in CBR involving the integration with IR systems, specifically where CBR has been effectively used to retrieve `most on-point' cases to guide the search and browse of vast IR collections~\cite{rissland1995hybrid}.
Our work draws inspiration from these integrative approaches, applying the principles of CBR to improve contextual understanding within LLMs.

LLMs are typically pre-trained on general text and subsequently fine-tuned on legal texts to learn domain-specific representations. 
Obtaining sufficiently large data sets for LLM training poses a significant challenge. 
For example, pre-trained LLMs may be fine-tuned using the LEDGAR dataset~\cite{tuggeneretal2020ledgar} for legal text classification downstream tasks or on extensive corpora like the Harvard Law case corpus\footnote{https://case.law/} using masked-language modelling or next-sentence prediction techniques.
Thereafter tested on other legal downstream tasks, such as those found in the CaseHOLD dataset~\cite{chalkidis2022lexglue}, which include tasks related to Overruling, Terms of Service, and CaseHOLD itself.
We observe that there remains a notable scarcity of LLMs specifically applied to legal question answering tasks. This is likely because existing legal QA datasets are mostly small, manually curated datasets (for example, the rule-qa task in the LegalBench collection~\cite{guha2023legalbench} is formed from 50 question-answer pairs). 
However the recent release of the Open Australian Legal Question-Answering (ALQA) dataset\footnote{\url{https://huggingface.co/datasets/umarbutler/open-australian-legal-qa}}, comprising over 2,100 question-answer-snippet triplets (synthesised by GPT-4 from the Open Australian Legal Corpus), presents an opportunity for LLMs to expand to legal QA.

\section{CBR-RAG: Using CBR to form context for LLMs} 
\label{sec:cbrrag}

In CBR-RAG, we integrate the initial retrieve-only stage of the CBR cycle with its indexing vocabulary and similarity knowledge containers to enable the retrieval of cases that serve as context for querying an LLM. 
Consequently, the original LLM query is augmented with content retrieved via CBR, creating a contextually enriched prompt for the LLM.

\begin{figure}[ht]
\centering    
\includegraphics[width=0.7\textwidth]{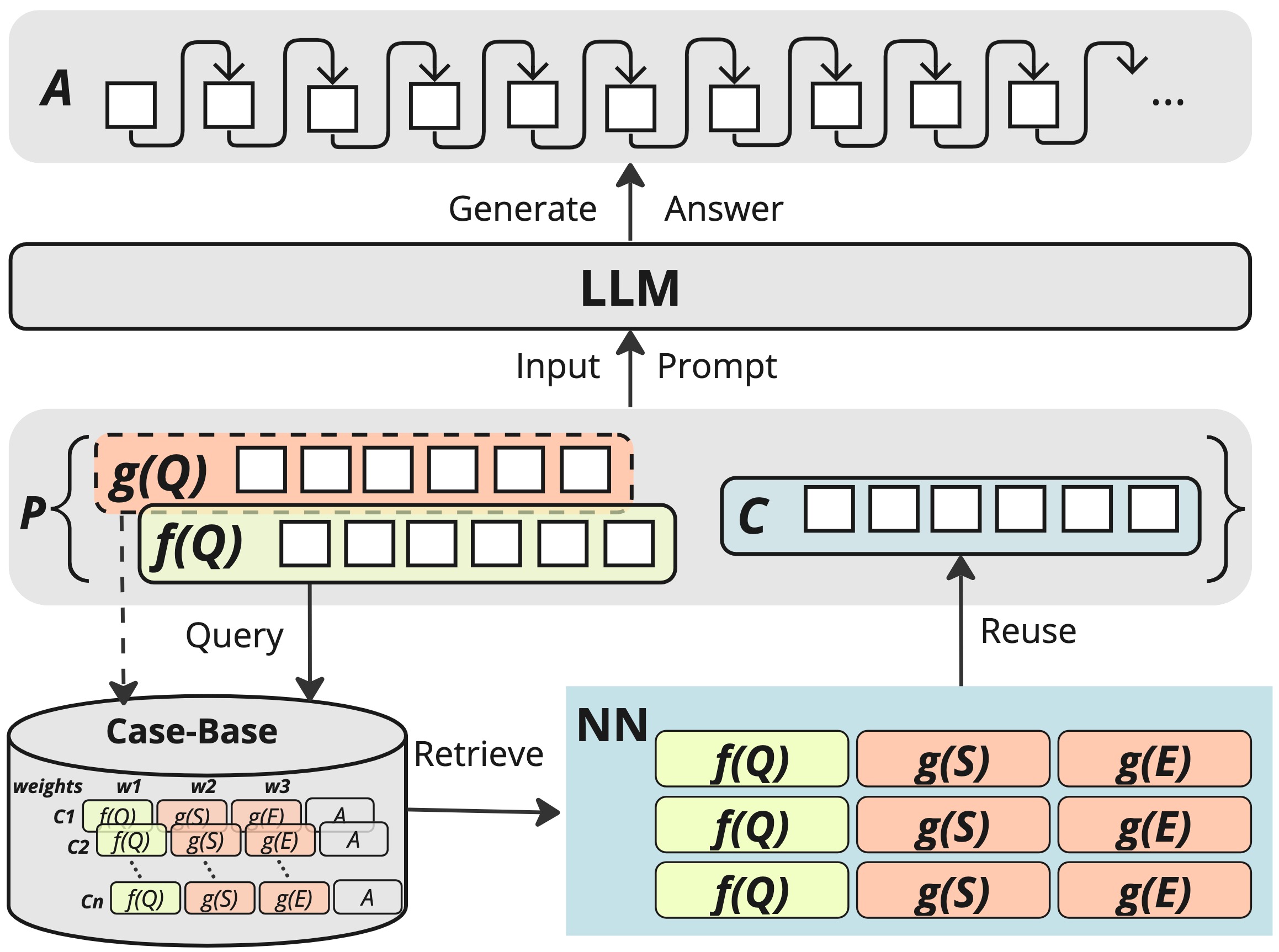}
\caption{CBR-RAG}
\label{fig:rag-cbr}
\end{figure}

Figures~\ref{fig:rag-cbr} illustrates a high-level architecture of a generative model for Question-Answering systems, highlighting the integration of CBR within it.
Here we denote the generative LLM model as, $G$, and the prompt, $P$, used to generate the response as a tuple, $p = (\mathcal{Q}, \mathcal{C} )$, where $\mathcal{Q}$ is the question reflecting the user's query, and, $\mathcal{C}$, is the context text with relevant details to guide the response generation. 
The response generated by the model for the query, $\mathcal{Q}$, is denoted as the answer, $A$.
%


\subsection{Casebase}\label{sec:casebase}
We used the Australian Open Legal QA (ALQA)~\cite{butler-2024-open-australian-legal-corpus} dataset to create the casebase. 
The dataset is formed of 2,124 LLM-generated question-answer pairs from the Australian Open Legal Corpus dataset. 
Each QA pair is corroborated by a supporting textual snippet from a legal document within the corpus for factual validation. 
An example case about the `interpretation of reasonable grounds in searches without a warrant' appears in Table~\ref{tab:case_example}.
Here the support text provides the context in which the question should be answered. 
The bold text further highlights examples of named entities that might usefully be captured separately for case comparison purposes.
\begin{table}[!h]
\centering
\renewcommand{\arraystretch}{1.2}
\caption{Examples Legal Q\&A case}
\label{tab:case_example}
\begin{tabular}{@{}p{0.2\linewidth}p{0.7\linewidth}@{}}
\toprule
\textbf{Component} & \textbf{Description} \\ \hline
Case Name & Smith v The State of New South Wales (2015) \\
Question & How did the case of Smith v The State of New South Wales (2015) clarify the interpretation of `reasonable grounds' for conducting a search without a warrant?  \\
Support  & In \CaseName{Smith v The State of NSW (2015)}, the plaintiff \Action{was searched without a warrant} \Location{near a known drug trafficking area} based on the plaintiff's nervous demeanor and presence in the area, but \Outcome{no drugs were found}. The legality of the search was contested, focusing on \LegalConcept{whether `reasonable grounds' existed}.\\ 
Answer & The case ruled `reasonable grounds' require clear, specific facts of criminal activity, not just location or behavior. 
\end{tabular}
\end{table}

\begin{figure}[!htb]
    \centering

    \begin{minipage}{0.45\textwidth}
        \centering
        \renewcommand{\arraystretch}{1.2}
        \begin{tabular}{| @{\hspace{5pt}} c @{\hspace{5pt}} | @{\hspace{5pt}}  p{0.9\linewidth}|}
        \hline
        \textbf{Index} &  \textbf{Act} (\textbf{Freq}) \\ \hline
        1 & Federal Court Rules (9) \\ \hline
        2 & Civil Aviation Regulations 1998 (8) \\ \hline
        3 & Land \& Env. Court Act 1979 (8) \\ \hline
        4 & Corporations Act 2001 (Cth) (7) \\ \hline
        5 & Migration Act 1958 (Cth) (6) \\ \hline
        6 & Customs Act 1901, Tariff Act 1995 (6) \\ \hline
        7 & Industrial Relations Act 1996 (6) \\ \hline
        8 & Environmental Planning and Assessment Act 1979 (5) \\ \hline
        9 & Trees (Disputes Between Neighbours) Act 2006 (5) \\ \hline
        10 & Migration Act 1958 (5) \\ \hline
        
        \end{tabular}
        \label{tab:indices_frequencies_acts}
    \end{minipage}
    \begin{minipage}{0.52\textwidth}
        \centering
        \adjustbox{angle=270,origin=c}{\includegraphics[width=1\textwidth]{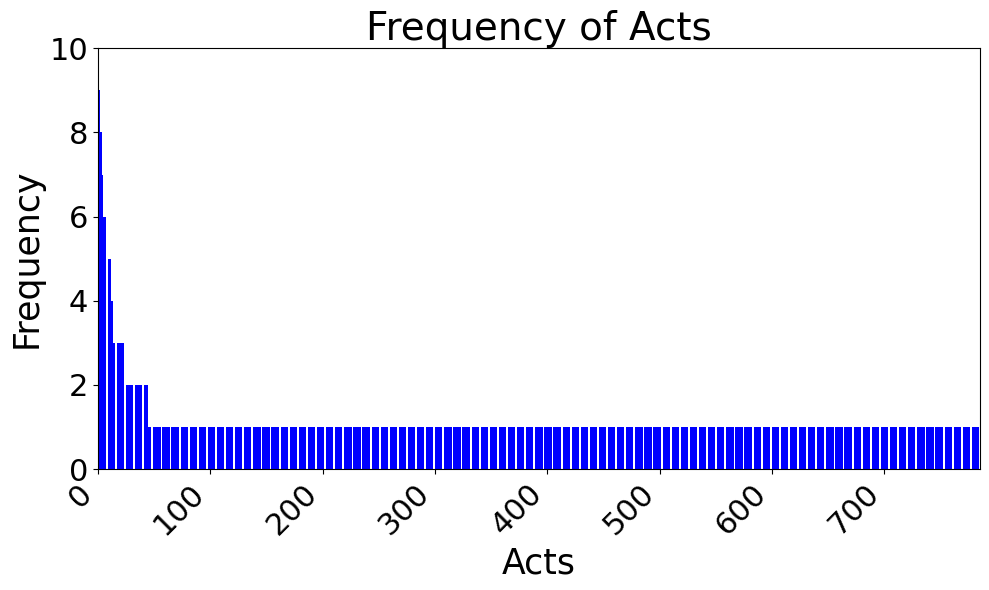}}
        \label{fig:case_act_frequency_distribution}
    \end{minipage}%
       \caption{ Ten most frequent legal acts in the casebase are listed on the left, and the legal act frequency distribution appears on the right.}
        \label{fig:qa_dataset}
\end{figure}
Figure~\ref{fig:qa_dataset} provides a frequency distributions of the legal acts identified in the casebase with the most frequently referenced legal acts in the dataset listed in the table (extracted using the prompt in Table~\ref{tab:scenarios-prompts1}). 
The most frequently mentioned act is the `Federal Court Rules' out of 785 unique legal acts. 
Within the dataset, 1,183 (57\%) cases were found to have no reference to legal acts, while only 44 acts appeared in more than 1 case. 
Accordingly, relying solely on legal acts for indexing would not be suitable for this casebase. 
Instead, it presents an ideal opportunity for experimentation with neural embeddings. 
These embeddings form the indexing vocabulary, with weighted similarity contributing to the similarity knowledge.

\section{Representation and Similarity} \label{sec:indexing}

We formalise our retrieval and representation methods. 
Let's denote the casebase as a collection of cases $C = \{c_1, c_2, \ldots, c_n\}$ containing question and answer legal cases.
Each case $c$, in the context of RAG is formalised as a tuple,
\[
c = \; < Q, S, E, A >
\]
where $Q$ represents a \textbf{question},  
$A$ represents the \textbf{answer}, 
$S$ represents the \textbf{support} for a given answer from an evidence base, and
$E$ represents a set of \textbf{entities} extracted from $S$. 
This case representation underpins RAG in its use of $S$. 
Typical CBR for question answering would be composed mainly of problem-solution components (i.e. question and answer respectively together with any lessons learnt), structuring cases as problem-solution-support enhances answer generation for the LLM by providing factually accurate context. 
Furthermore, while only the most relevant component of a document is extracted as the support text, the link to the full document is available in the original ALQA corpus. 

\subsection{Representation}
Initially, in textual form, each part is represented by neural embeddings. For diverse retrieval scenarios, we use a dual embedding form for $Q$ to enable matching it with not only other questions but where necessary matching it to the supporting text or the entities as follows: 
\begin{itemize}
    \item \textbf{Intra Embeddings}, $f(\cdot)$, optimised for attribute matching. These embeddings facilitate attribute-to-attribute comparisons where local similarities can be computed between the same types of attributes (e.g., questions with questions).
    \item \textbf{Inter Embeddings}, $g(\cdot)$, designed for information retrieval (IR) scenarios. These embeddings allow for matching that is not restricted to like attributes, enabling inter-attribute similarity assessment. This approach is particularly useful in situations where a question may be relevant for comparison to the support text or the entities.
\end{itemize}

A representation using intra-embedding is useful for tasks like semantic textual similarity that focus on finding sentences with closely related meanings, even if phrased differently. 
For example, a sentence like `The judge dismissed the case due to lack of evidence.' would find a semantically similar counterpart in `The court threw out the lawsuit because there was insufficient proof' 
Conversely, an inter-embedding representation is suited for tasks aimed at searching for documents relevant to a query. For example, a legal query on `copyright infringement in digital media' may yield cases with outcomes such as rulings on unauthorised content distribution or streaming without permission. 
These highlight the distinction 
between representations needed for ascertaining semantic similarity with intra-embeddings ($f(.)$), and 
finding relevant content with inter-embeddings ($g(.)$).

The dual-embedding case representation, accommodating both forms of retrieval tasks, is given by:
\[
c =\; < f(Q), g(Q), g(S), g(E), A >
\]
Similarly, the prompt can be expressed using the inter embedding as follows\footnote{Note: In our notation, we employ a calligraphic font for the prompt components ($f(\mathcal{Q}), g(\mathcal{Q}), \mathcal{C}$) to distinguish them from those of cases. Despite this stylistic difference, it is important to understand that both prompts and cases utilise similar embedding representations.}:
\[
p = \; < f(\mathcal{Q}), g(\mathcal{Q}), \mathcal{C} >
\]

\subsection{Case Retrieval} \label{sec:retrieve}
We use case retrieval to augment the context part of a prompt, $p$, given its query, $\mathcal{Q}$.
Accordingly, there are three comparison strategies for case retrieval: intra-, inter-, and hybrid-embedding based retrieval.
\\

\noindent
\textbf{Intra-embedding retrieval} involves matching on the basis of embeddings obtained from function $f(.)$. 
Here $f(\mathcal{Q})$ which represents the embeddings from the prompt's query is matched to query parts of the cases in the casebase. 
The best matching case identified from intra-embedding retrieval is defined by:
\begin{equation}
    \beta_k = \underset{c_i \in C, \, k}{\text{top-k}} \, \text{Sim}(f(\mathcal{Q}), f(Q_i))
    \label{eq:intra}
\end{equation}
Here, \text{top-k} refers to the selection of indices corresponding to the $k$ highest-scoring cases as determined by the similarity measure. 
$\beta$ represents the indices of these retrieved cases, 
while $Sim$ is a similarity metric (e.g., cosine similarity) that measures the similarity between the intra-embedding of the prompt's question and the intra-embeddings of the question parts of each case in the casebase.
\\

\noindent
\textbf{Inter-embedding retrieval} uses $g(\mathcal{Q})$ from the prompt to search the casebase, focusing on identifying relevant cases akin to an information retrieval style search, where attribute-to-attribute matching is not strictly followed. The best matching case is identified as follows:
\begin{equation}
    \beta_k = \underset{c_i \in C, \, k}{\text{top-k}} \, \; \text{Sim}(g(\mathcal{Q}), g(X_i))
    \label{eq:inter}
\end{equation}
where \(X_i\) can be either \(S_i\) (the Snippet part) or \(E_i\) (the Entity part) of the case.
\\

\noindent
\textbf{Hybrid embedding retrieval} is an alternative matching that involves a combination of both intra and inter-embeddings of the prompt's $\mathcal{Q}$ representations being used to match cases in a hybrid weighted retrieval approach:
\begin{equation}
\begin{split}
    \beta_k = \underset{c_i \in C, \, k}{\text{top-k}} \, (&w_1 \cdot \text{Sim}(f(\mathcal{Q}), f(Q_i)) + 
    w_2 \cdot \text{Sim}(g(\mathcal{Q}), g(S_i)) + \\
    & w_3 \cdot \text{Sim}(g(\mathcal{Q}), g(E_i)))
\end{split}
\label{eq:hybrid}
\end{equation}
This approach utilises both representation forms of the prompt's query $f(\mathcal{Q})$ and $g(\mathcal{Q})$ for retrieval.
Given the top $k$ cases $\{c_{\beta_j}\}_{j=1}^{k}$, we can extract the context for the prompt based on the retrieval option $\rho$ as follows:

\begin{subnumcases}{\text{Context,} \; \mathcal{C}(\rho) =}
    \{S_{\beta_j}\}_{j=1}^{k} & if $\rho = 1$, ``support-text-only'' \label{eq:support}\\
    \{Q_{\beta_j}, S_{\beta_j}, E_{\beta_j}, A_{\beta_j}\}_{j=1}^{k} & if $\rho = 2$, ``full-case'' \label{eq:case}
\end{subnumcases}


\section{Embedding models}\label{sec:encoders}
\begin{figure}[ht]
\centering    
\includegraphics[width=0.75\textwidth]{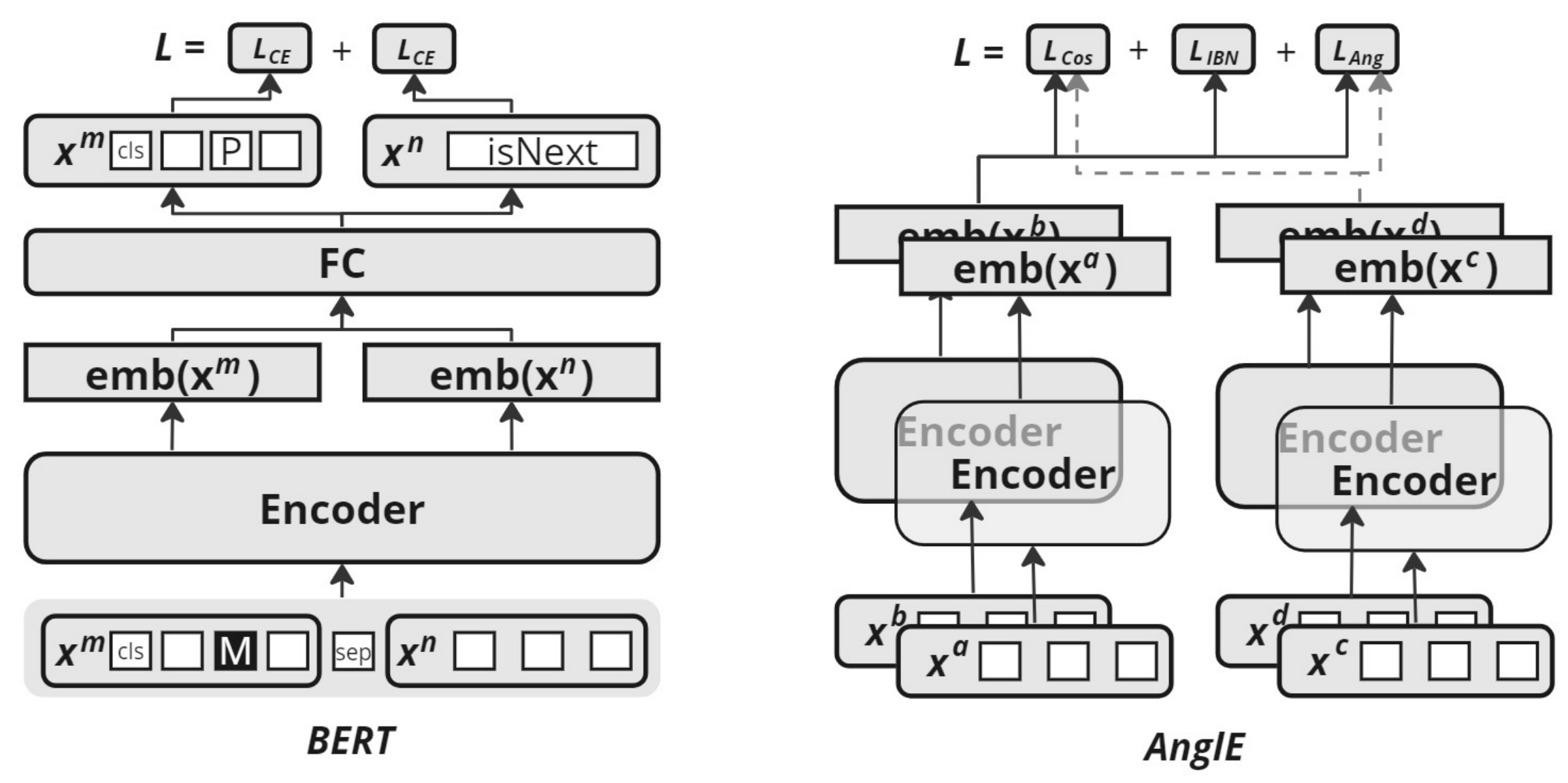}
\caption{Architecture and training process for \bert and \abert. Note that \lbert has the same architecture as \bert, but is pre-trained on legal text.}
\label{fig:encoders}
\end{figure}

In this work, we explore embeddings generated by \bert, \abert, and \lbert; the latter being pre-trained on diverse English legal documents, with the rest all being general-purpose embeddings.
We next provide an overview of these models, as illustrated in 
Figure~\ref{fig:encoders}, 
and discuss how they can be used to generate the $f$
and $g$ forms of representations.

\subsection{\bert}
The Bidirectional Encoder Representations from Transformers (\bert) model~\cite{devlin2018bert} is a language model for learning generalisable text embeddings. The model is formed of an encoder block (taken from the transformer architecture in~\cite{vaswani2017attention}), followed by a fully-connected layer. 
The bidirectional nature of \bert is derived from its pre-training technique, which conditions on both the left and right contexts of input sentences simultaneously.
The model uses a self-supervised learning strategy that combines masked-language modeling (MLM) with next sentence prediction (NSP) to acquire contextually rich word embeddings. During training, a pair of sentences is fed into the model, and a random subset of words is replaced with the [MASK] token, establishing a sequence-to-point task. 
The objective here is to predict the masked words by using the context provided by both previous and subsequent words. 
Furthermore, a [CLS] token is inserted at the beginning of the input sequence to accumulate contextual information from the entire sequence, facilitating tasks like sentence relationship classification. 
Pre-training with both MLM and NSP, can be seen as typical text prediction and is controlled by the standard cross-entropy objective function. 
\bert is pre-trained on a combination of large general purpose text datasets totalling 3.3B words and has demonstrated strong performance across many domains.

 \subsection{\lbert Trained on General Legal Data} \label{lbert}
Domain-specific knowledge is known to be beneficial for legal tasks~\cite{chalkidis2020legalbert,tuggener-etal-2020-ledgar}. For example, the word `case' may refer to a variety of containers (brief case, suit case, display case, etc), but also a court case. While semantic relations with the latter are likely to be impactful for legal question answering, the greater frequency of the former in general corpora will result in embeddings more weighted towards that context. The \lbert family of models~\cite{chalkidis2020legalbert} enhance the \bert model by further pre-training on an additional 
12 GB of diverse English legal 
documents from a mix of UK, EU, and US legislation and case law. The goal is to embed domain knowledge into produced embeddings by learning from a relevant document set.

\subsection{\bert with AnglE Embeddings}

The AnglE embeddings~\cite{li2023angleoptimized} adopts a contrastive learning strategy (similar to Siamese networks~\cite{bromley1993signature}) to learn embeddings through matching both positive and negative text pairs, where a positive pair is considered to be similar (above some threshold usually). 
The process begins with \bert embeddings, input to AnglE for optimisation.
Its novelty stems from one of the three loss functions, that are designed to overcome the vanishing gradient issue encountered in cosine similarity-based comparisons, particularly at the extremes of similarity (or dissimilarity). This is achieved by comparing embeddings based on angle and magnitude within complex spaces, effectively bypassing cosine similarity's saturation problem. Once trained the AnglE embedding model can be used to generate text embeddings using a final pooling layer~\cite{li2023angleoptimized}.

One of the difficulties when using \abert for a specific domain is that one must generate a supervised training set (unlike with \bert which can be trained in an unsupervised manner using the MLM and NSP self-supervision methods). 
This is because the AnglE method adopts a contrastive learning strategy, where the supervised dataset must include paired instances for training. This can be prohibitive in contexts where domain expertise is required for labelling. 

\subsection{Dual-embedding Case Representation with AnglE}
For the purposes of this work, we leverage dual-embeddings introduced in~\cite{li2023angleoptimized}.  
In terms of the inter-embedding retrieval,  a specific embedding prompt cue
$Cue(\mathcal{Q})$
is used to contextualise the relevance of the query embeddings towards matching with attributes other than that of questions, as follows:
\[
Cue(Q) = \text{``Represent this sentence for searching relevant passages: ''} \{Q\}
\]
The idea here is that the cue is used to influence the embedding generation towards inter-retrieval oriented embeddings as follows: \[g(Cue(\mathcal{Q}))\]
For intra-embedding retrieval the prompt text is empty, i.e. input the text without specifying an additional prompt cue. 
This means the embedding function $f$ processes the query $\mathcal{Q}$ directly, as $f(\mathcal{Q})$. Table~\ref{tab:questions_comparison} provides an example of question and support text version that are used as input to each of \bert, \abert and \lbert with relevant prompts to enable the generation of the alternative embeddings that we intend using during case matching with CBR.
\begin{table}[!htb]
\centering
\renewcommand{\arraystretch}{1.2}
\caption{Comparison of an example question with and without the $Cue$ text (in blue) to create inter and intra embeddings.}
\begin{tabular}{p{0.1\linewidth} p{0.8\linewidth}}
\hline
\multicolumn{1}{l}{\textbf{Embedding}} & \textbf{Question} \\ \hline
\multirow{2}{*}{intra} & $f$(``What were the court's findings regarding the financial liabilities of Allco Finance Group Ltd to Blairgowrie Trading Ltd?") \\
\hline
\multirow{3}{*}{inter} & $g$(\textcolor{blue}{``Represent this sentence for searching relevant passages:"} + ``What were the court's findings regarding the financial liabilities of Allco Finance Group Ltd to Blairgowrie Trading Ltd?" )\\
\hline
\end{tabular}
\label{tab:questions_comparison}
\end{table}

\section{Evaluation} \label{sec:eval}

The aim of our evaluation is two-fold: 
1) to understand the impact of inter and intra embeddings on weighted retrieval, and 2) to understand the generative quality of RAG systems when coupled with a case-based retrieval-only system. 
To analyse weighted retrieval we compared several representation combinations of embedding models and similarity weights as follows:
\begin{itemize}
    \item compare three alternative forms of text embeddings using the encoders presented in  Section~\ref{sec:encoders} - \bert, \lbert and \abert;
    \item compare four alternative weighting schemes to assess utility of question, support, and entity components within case representations. These include:
    \begin{itemize}
    \item Question only, represented by the weights [1,0,0] (see Equation~\ref{eq:intra}),
    \item Support only, with weights [0,1,0],
    \item Entities only, using weights [0,0,1] (see Equation~\ref{eq:inter}),
    \item A hybrid approach, combining these components with weights [0.25, 0.40, 0.35] (refer to Equation~\ref{eq:hybrid}).
    \end{itemize} 
\end{itemize}
Accordingly \bert{[0,1,0]} would denote using a Support only version of retrieval with the \bert embedding; and using the same naming convention, \abert{[0.25,0.4,0.35]} indicates a hybrid dual embedding method where \abert embeddings are used with the specified weights for case retrieval. 
Our baseline comparator is an LLM with no case retrieval i.e. No-RAG.

We selected Mistral~\cite{jiang2023mistral} for answer generation at test time, due to its open-source availability, allowing us to use a model distinct from OpenAI's GPT-4, which was employed for formulating the Q\&A casebase. This approach effectively simulates consulting an alternative expert in place of a human specialist.



\subsection{Legal QA Dataset Analysis}\label{subsec:datavalidate}
The ALQA dataset, introduced in Section~\ref{sec:cbrrag} is a synthetic Q\&A dataset generated from real legal documents in the Australian Open Legal Corpus. 
Here sentences extracted from documents, each coupled with a prompt, are used to generate corresponding questions and answers from OpenAI's GPT-4 model. 
We performed a multi-stage analysis to ensure that this dataset was appropriate for our retrieval and follow-on answer generation tasks. 
\begin{figure}[ht]
\centering
\includegraphics[width=0.85\textwidth]{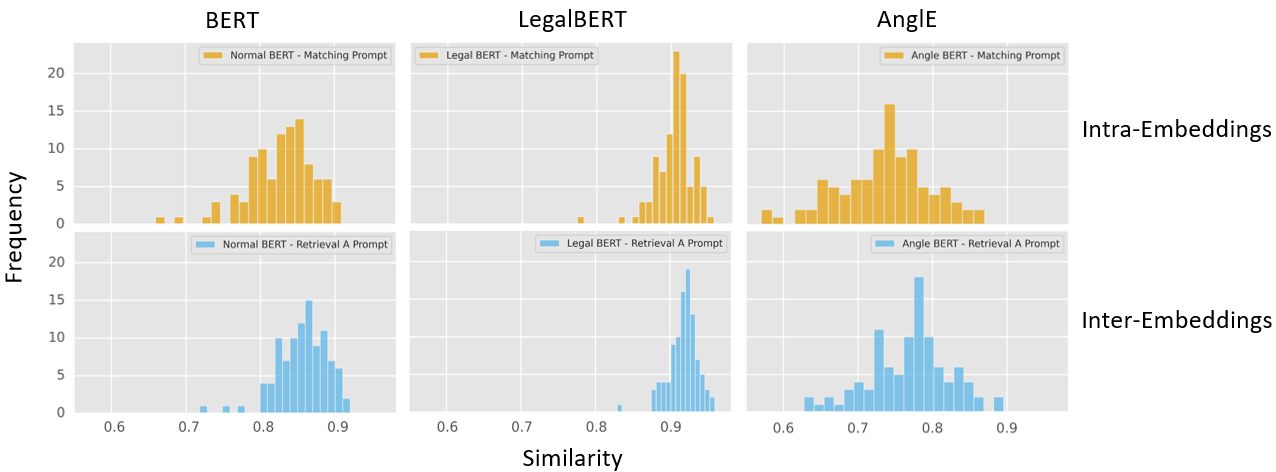}
\caption{Cosine similarity distribution for intra- and inter-embeddings.}
\label{fig:sim_embs}
\end{figure}

\textbf{Dataset Validation} 
involved a randomly sampled manual analysis of the questions and answers performed by the research team to ensure the dataset contained no LLM-based anomalies (i.e. hallucination, factually incorrect statements, etc). 
We then converted the question text into intra- and inter-embeddings using \bert, \lbert and \abert and examined the similarity distribution by calculating the cosine similarity between each instance and its nearest neighbour (shown in Figure~\ref{fig:sim_embs}). We observe that the similarity distributions are mostly Gaussian, with the most frequent values between 0.7 and 0.8 cosine similarity for both embedding types produced by \bert and \abert. 
Embeddings learned by \lbert seem to be more densely clustered, as indicated by the higher similarity.
This could suggest a reduced ability to discriminate based on similarity compared to \bert and \abert. 
We believe the results of this analysis were very promising, as they suggest the embeddings avoid the traditional issues associated with feature engineering approaches (such as sparse similarity distributions).

\textbf{Case-Base}
consisted of the ALQA dataset where each case consists of the full Q\&A content as discussed in Section~\ref{sec:casebase}. While originally containing 2,124 question-support-answer triplets, 40 were removed due to offensive content, and therefore our case-base contains 2,084 cases.
The case representation was also expanded to include entities to form the complete tuple as discussed in Section~\ref{sec:indexing}.

\begin{table}[htbp]
\centering
\renewcommand{\arraystretch}{1.2}
\caption{Prompts used in this research.}
\begin{tabular}{@{}p{0.33\linewidth} @{\hspace{3pt}} p{0.66\linewidth}@{}}
\toprule
\textbf{Scenarios \& Generator} & \textbf{Prompt} \\
\midrule
Extract legal acts to pair cases for test set creation. \newline
\textbf{Gpt-3.5-turbo-0125} & Extract the legal act(s) in this text. Print `None' if nothing is found. \{ TEXT \} \\ 
\midrule
Extract entities for case representation. \newline \textbf{Gpt-3.5-turbo-0125} & Extract named entities and unique identifiers as a single text (separated with white-space) line from this " + TEXT. Print `' if nothing is found. \{ TEXT \} \\ 
\midrule
Generate Qs from text pairs for test set creation. \newline \textbf{Mistral-7B} & Produce a question and answer where the answer requires detailed access to both Text 1 and Text 2. Don’t refer texts in the question or answer. \{ Text1: TEXT1 — Text2: TEXT \} \\ 
\midrule
Generate answers from snippets for retrieval analysis. \newline \textbf{Mistral-7B} & Answer QUESTION by using the following contexts: \{ TEXT1 — TEXT2 \} \\ 
\midrule
Generating answers from cases for retrieval analysis. \newline \textbf{Mistral-7B} & Answer QUESTION as a simple string (with no structure) by using the following question, citation, and answer tuples as context: \{ Question: Q1, Citation: C1, Answer: A1 — Question: Q2, Citation: C2, Answer: A2 \} \\
\bottomrule
\end{tabular}
\label{tab:scenarios-prompts1}
\end{table}

\textbf{Test Set Creation}
focused on creating a discrete test set of questions that reference applicable knowledge in the case-base, without directly mapping to a single case in the case-base. 
To guide the generation of unique questions for test purposes, we first analysed the case-base in terms of unique legal acts mentioned in all cases. 
We then selected case pairs based on the common acts and, using Mistral-7B~\cite{jiang2023mistral} generated 35 new question-answer pairs, each answerable using the combined information from both cases in a pair, as detailed in the prompt presented in Table~\ref{tab:scenarios-prompts1}. 
The rationale is that by pairing cases with common legal acts, we can encourage Mistral to create novel test Q\&A pairs. These pairs are unique in that they necessitate synthesising information from both cases in the pair to form a coherent question and corresponding answer, ensuring that the resulting pairs differ from the questions and answers of the individual cases in the case-base, thereby creating a set of new test cases that are reasonably disjoint.
All outputs were manually reviewed to ensure they were suitable test cases\footnote{Test dataset available at \href{https://huggingface.co/datasets/Ramitha/open-australian-legal-qa-test}{open-australian-legal-qa-test}}.
After pruning cases that lacked suitable answers or contained answers merely linked to the case pair by conjunctions, we were left with a total of 32 cases.

\subsection{Retrieval Analysis}
\begin{figure}[ht]
\centering
\includegraphics[width=0.85\textwidth]{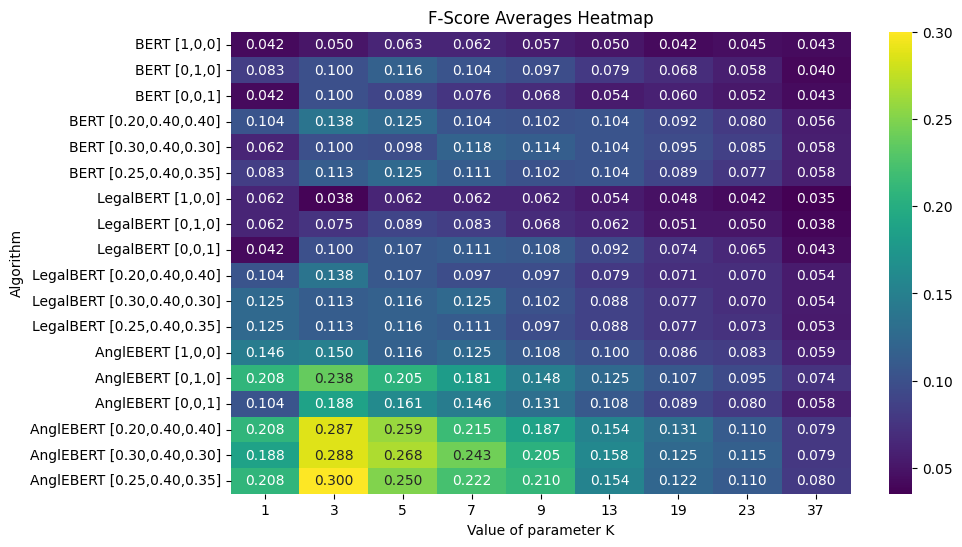}
\caption{F1 score for Retrieval@k}
\label{fig:f1-score_heatmap_against_retrieval_at_k}
\end{figure}

We first evaluated the quality of case retrieval, by exploiting the fact that each of our 32 test cases had originated from a pair of parent cases. 
Accordingly, for each test case, we treated the pair of parent cases from the training set as relevant, and all other cases as irrelevant (allowing calculation of ranked precision and recall). 
We then performed a similarity-based retrieval using the k-Nearest Neighbors algorithm, exploring a range of $k$ values consisting of prime numbers between 1 and 37
We calculated results using F1-score for retrieval@k and visualised using a heat-map (see Figure~\ref{fig:f1-score_heatmap_against_retrieval_at_k}). 
Here the best performing algorithm was Hybrid \abert with [0.25, 0.40, 0.35] weights and $k = 3$. 
Accordingly, $k$ was selected as 3 to be used in the subsequent generative experiments with $k=1$ as a comparative baseline.


\subsection{Generation Results}

We evaluated the quality of generated output using six algorithms run with $k = 1$ and $k = 3$, along with the baseline for No-RAG. The results are presented in Table 4.
\begin{table}[h]
\centering
\caption{Cosine scores for hybrid algorithms}
\label{my-label}
\begin{tabular}{clccc}
\toprule
& &\textit{No Context} & \textit{Support} & \textit{Full Case} \\
\cmidrule{1-5}
\multirow{6}[-25]{*}{\( k = 0 \)} & {No-RAG} & 0.8967 &
\\
\midrule
\multirow{6}[-9]{*}{\( k = 1 \)} & \multirow{2}[-5]{*}{Hybrid \bert} &-& 0.8986 & \textbf{0.9068} \\

& \multirow{2}[-5]{*}{Hybrid \lbert} &-& 0.9020 & 0.9043 \\

& \multirow{2}[-5]{*}{Hybrid \abert} &-& 0.9121 & 0.9074 \\
\midrule
\multirow{6}[-9]{*}{\( k = 3 \)} & \multirow{2}[-5]{*}{Hybrid \bert} && 0.9007 & 0.8998 \\

& \multirow{2}[-5]{*}{Hybrid \lbert} &-& 0.9034 & \textbf{0.9045} \\

& \multirow{2}[-5]{*}{Hybrid \abert} &-& 0.9092 & \textbf{*0.9141} \\

\bottomrule
\end{tabular}
\end{table}

Next, we evaluated the quality of generated output. Six algorithms were run with $k = 1$ and $k = 3$, along with the baseline for No-RAG.
 
The Mistral-7B-open model was utilised as the LLM to generate the answers given a question from a test case with the RAG context (formed using the CBR-RAG retrieval approaches). 
The generated content was then converted to an embedding using Mistral for cosine-based comparison with the expected answer from the test case (i.e. reference text) for comparison, with the expectation that higher the similarity the better the CBR-RAG setup. 

Relatively high cosine scores were observed with the No-RAG baseline due to the generator having parametric memory in most of the legal questions asked by default. The best semantic similarity was noted with the Hybrid \abert variant when 3 nearest neighbours were fed into the generator in the form of full cases forming context for RAG. 
It provided answers on average with 1.94\% increase in performance. All hybrid variants performed better than the No-RAG baseline. 
We also observed that including the full case in the prompt provided better results compared to including only the Support text in most hybrid algorithms.
Overall, Hybrid \abert outperforms the \bert and \lbert variants with higher semantic similarity observed when $k=3$.


We performed a series of ANOVA tests to evaluate whether results were significant.
Following this we carried out paired tests between the best-performing methods from each of the \bert and \lbert groups (in bold), as well as the baseline No-RAG. 
Here we found that hybrid \abert k=3, significantly outperforms (asterik) both 'No-RAG' and hybrid \lbert k=3, at the 95\% confidence level, as shown by a one-tailed T-test. Against hybrid \bert k=1, \abert k=3 shows significant improvement at the 90\% confidence level.


 

\section{Conclusions}
\label{sec:conc}
In this paper we have presented CBR-RAG, improving LLM output by augmenting input with supporting information from a case-base of previous examples. 
We have performed an empirical evaluation of different retrieval methods with knowledge representation and comparison using \bert, \lbert, and \abert embeddings. 
Responses generated by CBR-RAG outperforms those of baseline models in similarity to ground truth. 
Our findings confirm that using a case-retrieval approach in RAG systems lead to clear performance benefits, but selecting an appropriate embedding for case representation is key. 
A qualitative analysis with a domain expert would be an ideal next step to validate these results. 
%
The fact that \abert had the best performance suggests that its contrastive approach to optimising for embeddings (based on similarity comparisons) remains more important than the standard self-supervised masked training strategies used by \lbert, even if trained on general legal data.  
We note that none of the embedding methods, including \lbert (trained on broader legal collections), were fine-tuned to the ALQA-specific legal corpus.
We are keen to explore the impact of fine-tuning using contrastive self-supervision methods and determine the necessary data supervision burdens for this process, which could pose disadvantages in certain domains.

In future work, we aim to explore further opportunities within representation and specifically alternative methods for text embeddings. 
Moreover, given the hybrid embeddings' success, which combines multiple representations for fine-grained similarity comparison, we are keen to expand CBR-RAG with more retrieval capabilities. 
Finally, we found that combining multiple neighbours while maintaining a coherent prompt is challenging, so we plan to explore case aggregation strategies in the future.
\bibliographystyle{splncs03}
\bibliography{ref}

\end{document}